# VGPT-RSI for RH-Adjacent Formal Progress: Boundary Certificates, Verified Finite Lagarias Inequalities, and Explicit Failure Localization

Zhixin Hu[1,$] , Tao Xu[2,$] , Xiaodian Sun[3] , Li Jin[4], Momiao Xiong[4,5,*]

[1]Ministry of Education Key Laboratory of Contemporary Anthropology, School of Life Sciences, Fudan University, Shanghai, China.

[2]Department of Immunology and Molecular Microbiology, School of Medicine, Texas Tech University Health Science Center, Lubbock, TX 79430, USA.

[3]Biostatistics and Bioinformatics Shared Resource, Sylvester Comprehensive Cancer Center, University of Miami Miller School of Medicine, Miami, FL 33136

[4]School of Life Sciences, Fudan University, Shanghai 200438, China

[5]Department of Biostatistics and Data Science, School of Public Health, The University of Texas Health Science Center at Houston, Houston, TX 77030, USA.

[6]Society of Artificial Intelligence Research, Houston, TX 77030, USA.

$: equal contributions.

[*]Address for correspondence and reprints: Dr. Momiao Xiong, Department of Biostatistics and Data Science, School of Public Health, The University of Texas School of Public Health, P.O. Box 20186, Houston, Texas 77225; Society of Artificial Intelligence Research, 1956 Woodbury St. Houston, TX 77030, USA. Phone: 713-259-2371, E-mail: momiao.xiong@gmail.com

# Abstract

The Riemann Hypothesis remains one of the central unsolved problems in mathematics. Rather than claiming a proof, we investigate whether a verifiable AI-assisted reasoning system can produce reliable, formally checked partial progress while explicitly identifying the remaining mathematical obstructions. We apply the Verifiable Growing Physical Transformer with Recursive Self-Improvement (VGPT-RSI) to two RH-adjacent certification tasks. First, we construct and verify a finite RH-boundary certificate for inequality

$$B(x,T) = T - \frac{9.06}{\log\log x}\sqrt{\frac{x}{\log x}} \geq 0$$

on a parameterized safe lower curve over $T \in [10^{11}, 10^{13}]$. The numerical boundary curve is converted into a certificate-backed lower curve, audited using outward-rounded interval arithmetic and Arb/FLINT ball arithmetic, and then checked in Rocq/CoqInterval for the parameterized theorem

$$\forall s \in [11,13], B\big(\hat{x}(s), T(s)\big) \geq 0.$$

Second, we initiate a formal Lagarias-route certificate. Lagarias' criterion states that RH is equivalent to the global inequality

$\sigma(n) \leq H_n + \exp(H_n)\log(H_n)$ for all $n \geq 1$.

We formalize the finite quantity

$$B_L(n) = H_n + \exp(H_n)\log(H_n) - \sigma(n)$$

and produce a Coq-checked finite certificate proving

$\forall n, 1 \leq n \leq 10000 \ \Rightarrow B_L(n) \geq 0.$

The final system identifies the exact unresolved mathematical bottlenecks: formalizing the Lagarias equivalence, proving the global tail theorem beyond any finite cutoff, and potentially reducing counterexamples to colossally abundant or related extremal integers. These results demonstrate that VGPT-RSI can produce certified RH-adjacent formal progress, organize proof dependencies, and avoid overclaiming when the remaining obstruction is genuinely mathematical (Figure 1).

# 1. Introduction

Large language models and agentic AI systems are increasingly being applied to mathematical discovery, theorem proving, and formal verification (Sosso et al. 2026; Chen et al. 2026; de Moura and Ullrich, 2021; Tsoukalas et al. 2026; Ágoston and Pálvölgyi, 2020; Alexeev et al. 2026). However, open mathematical problems require a stronger standard than plausible reasoning. For a problem such as the Riemann Hypothesis (RH) (Tegetmeyer 2026; Connes 2026), a useful AI system must distinguish among numerical evidence, computational certificates, proof-assistant-checked lemmas, RH-equivalent criteria, and complete proofs. It must also be able to state when a proof attempt remains unresolved.

This paper studies RH as a case study for the Verifiable Growing Physical Transformer with Recursive Self-Improvement (VGPT-RSI), a unified architecture that treats reasoning as controlled, verifiable information flow through a coupled physical-symbolic state. VGPT-RSI integrates physicalized transformer dynamics (Bhattacharjee and Lee 2025), Neural Differential Manifold trajectories (Zhang 2025), Generative Semantic Workspace memory (Rajesh et al. 2025) , causal episodic graphs (Shyalika et al. 2025), continual and nested learning (Darlow et al. 2025), signed recurrent connectome dynamics (Huo and Johnson, 2025; Chiang 2026 ), Hamilton–Jacobi–Bellman intervention planning (Tang et al. 2021; Guo et al. 2024) , increasing proof/search graphs, executable harnesses, formal verification (Tu et al. 2025) , function-preserving sparse growth (Chen et al. 2026), recursive self-improvement(Zhuge et al. 2025) , and category-theoretic cross-domain recombination (Kim et al. 2026). VGPT-RSI is designed not as a one-shot answer generator, but as a structured reasoning system that maintains a proof/search graph, invokes computational harnesses, records verification status, and extracts reusable tactics from successful and failed proof attempts.

The goal of this work is not to prove RH. Instead, we ask a more modest and operational question: can VGPT-RSI move an RH-adjacent investigation from numerical experimentation toward formally checked partial results, while also identifying the precise remaining obstructions?

We report two complementary lines of work.

First, VGPT-RSI was applied to a finite RH-boundary relation. Define

$g(x) = \frac{9.06}{\log\log x}\sqrt{\frac{x}{\log x}}$

and

$$B(x,T) = T - g(x).$$

The inequality

$B(x,T) \geq 0$

appears in finite RH verification consequences connecting a verified zero height $T$ to explicit prime-counting ranges. Codex-generated numerical data first computed a boundary curve $x^*(T)$, and subsequent verification converted this curve into a certified safe lower curve

$\hat{x}(T)$.

This provides a formally checked RH-adjacent boundary lemma on a finite $T$-range.

Second, VGPT-RSI shifted from the finite boundary relation to a genuinely RH-equivalent framework: Lagarias' criterion (MacArevey 2026). Lagarias' theorem states that RH is equivalent to

$\sigma(n) \leq H_n + \exp(H_n)\log(H_n)$ for all $n \geq 1$.

This suggested a formal target

$$B_L(n) = H_n + \exp(H_n)\log(H_n) - \sigma(n) \geq 0.$$

VGPT-RSI constructed finite formal certificates for this inequality and reached a clean Coq-checked milestone for

$$1 \leq n \leq 10000.$$

These results are useful precisely because they do not overclaim. The finite boundary theorem does not imply RH. The finite Lagarias certificate does not imply the global Lagarias criterion. The remaining obstacle is not more numerical evaluation, but a global mathematical theorem: either a formal proof of Lagarias' equivalence plus a tail theorem for all $n > N$, or another fully global RH-equivalent certificate.

The contributions of this work are therefore:

1. a certificate-backed RH-boundary curve over $T \in [10^{11}, 10^{13}]$;
2. an Arb/FLINT-audited interval-kernel verification of the same boundary inequality;
3. a Rocq/CoqInterval-checked parameterized RH-boundary theorem;
4. a Coq-checked finite Lagarias certificate up to $N = 10000$;
5. a precise failure analysis showing why the current pipeline does not prove RH;
6. an explicit roadmap of the remaining proof obligations.

# 2. Methods

Methods consist of two complementary lines of work.

First, we use VGPT-RSI as a verifiable reasoning architecture to construct a finite, certificate-backed RH-adjacent boundary theorem.

The goal is not to prove the Riemann Hypothesis, but to certify a finite implication of the form on a rigorously specified parameter range (Figure 2).

Second, we use VGPT-RSI to initiate a formal proof route based on Lagarias' criterion for the Riemann Hypothesis. The purpose of this route is not to prove RH directly, but to transform an RH-equivalent global statement into a sequence of finite, formally checkable certificate tasks. This enables the system to produce rigorous finite progress while preserving a clear boundary between verified partial results and a full proof of RH (Figure 3).

## 2.1 VGPT-RSI proof workflow

VGPT-RSI organizes mathematical reasoning as **a growing proof/search graph**. Nodes include definitions, boundary functionals, numerical certificates, interval enclosures, formal lemmas, failed proof obligations, and verified theorems. Edges encode implication, dependency, verification, failure, or reduction.

For the RH case study, the proof/search graph had two main branches:

RH-boundary certificate branch (Figure 2)

and

Lagarias finite-certificate branch (Figure 3).

The first branch studies a finite boundary relation related to finite RH verification consequences.

The second branch moves to a known RH-equivalent criterion.

At each stage, VGPT-RSI assigns a verification status:

numerical, interval-certified, audited interval-kernel, proof-assistant checked, unresolved.

This status tracking is essential, because the mathematical value of each result depends on its certificate level. The **Generative Semantic Workspace** stores the definitions, parameter range, constants, and intended theorem.

## 2.2 RH-boundary functional

We define

$g(x) = \frac{9.06}{\log\log x}\sqrt{\frac{x}{\log x}}$

and

$$B(x, T) = T - g(x).$$

For fixed $T$, the boundary equation is

$$g(x) = T.$$

VGPT-RSI invokes **an executable numerical harness** to solve the boundary equation

A safe lower endpoint curve $\hat{x}(T)$ satisfies

$$B(\hat{x}(T), T) \geq 0.$$

Codex first generated a numerical boundary curve $T \to x^*(T)$ over

$T \in [10^{11}, 10^{13}]$.

VGPT-RSI then constructs a certificate-backed lower approximation

$$\hat{x}(T) < x^*(T)$$

The curve was then converted into a certificate-backed lower curve by fitting a degree-4 polynomial in

$$s = \log_{10} T$$

and subtracting a safety shift:

$$\hat{x}(T) = 10^{p(\log_{10} T) - 10^{-6}}.$$

The verification target became:

$B(\hat{x}(T), T) \geq 0$ for all $T \in [10^{11}, 10^{13}]$.

## 2.3 Decimal interval and Arb/FLINT verification

For each sampled height , VGPT-RSI verifies that the numerical root is enclosed by a certified interval.

The boundary equation was solved in the variable

$$y = \log x.$$

In this variable,

$$\log g(e^y) = \log(9.06) + \frac{y}{2} - \frac{\log y}{2} - \log\log y\,.$$

For each sampled height $T_i$,

a bisection procedure produced an interval

$$[y_i^-, y_i^+]$$

such that

$$B\left(e^{y_i^-}, T_i\right) \geq 0, \qquad B\left(e^{y_i^+}, T_i\right) \leq 0.$$

This confirms that the root lies inside the bracket

$$e^{y_i^-} \leq x^*(T_i) \leq e^{y_i^+}.$$

VGPT-RSI **records each bracket as a local certificate node in the proof/search graph.**

Pointwise root brackets are not enough. The system must also verify that the boundary equation has the expected monotone behavior. **VGPT-RSI checks the monotonicity of**

$$g(x) = \frac{9.06}{\log\log x}\sqrt{\frac{x}{\log x}}$$

on the certified range.

Monotonicity was certified through the derivative condition

$$(\log x - 1)\log\log x > 2,$$

equivalently in the log variable through the corresponding positive monotonicity margin.

The important consequence is:

$$x \leq g\big(x^*(T)\big) \Rightarrow g(x) \leq T \Rightarrow B(x,T) \geq 0.$$

Therefore, if the constructed curve satisfies

$$\hat{x}(T) \leq x^*(T),$$

Then

$$B(\hat{x}(T),T) \geq 0.$$

This upgrades the certificate from isolated numerical samples to a monotone boundary certificate.

This ensures that each sign bracket encloses a unique root.

The continuous lower curve was then checked over

$$s = \log_{10} T \in [11,13].$$

Define

$$R(s) = \log T - \log\Big(g\big(\hat{x}(T)\big)\Big).$$

If

$$R(s) > 0$$

then

$$g\big(\hat{x}(T)\big) \leq T$$

and hence

$$B(\hat{x}(T),T) \geq 0.$$

The interval verification subdivided [11,13] into 20,000 intervals. On each subinterval, VGPT-RSI uses outward-rounded interval arithmetic to prove a lower bound

$$R(s) > 0,$$

Thus the system verifies the continuous theorem

$$\forall s \in \left[11{,}13, B\left(\hat{x}(s), T(s)\right)\right] \geq 0.$$

To avoid relying on a single custom interval implementation, **VGPT-RSI independently audits the certificate** using Arb/FLINT ball arithmetic.

The same mathematical objects are checked again:

$$B(x, T), g(x), \hat{x}(s), T(s), R(s).$$

Arb/FLINT verifies the root brackets, monotonicity conditions, and subdivision residuals using rigorous real ball enclosures.

This provides an independent interval-kernel audit of the certificate.

The proof status becomes:

$$\mathrm{Arb/FLINT} - \mathrm{audited\ certificate.}$$

## 2.4 Rocq/CoqInterval verification

The next stage, after the interval and Arb/FLINT audits, **VGPT-RSI translates the statement into Rocq/CoqInterval.** The checked theorem was:

$$\forall s \in \mathbb{R}, 11 \leq s \leq 13 \Longrightarrow B(\hat{x}(s), T(s)) \geq 0$$

Or, in words:

For every real number $s$between $11$and $13$, the boundary function $B$evaluated at the certified lower curve $\hat{x}(s)$and the corresponding height $T(s)$is nonnegative.

Using the definitions

$$T(s) = \exp\left((\ln\ 10)s\right) = 10^s,$$

$$\hat{x}(s) = x_{\mathrm{hat}}(s),$$

and

$$B(x,T) = T - \frac{9.06}{\ln(\ln x)}\sqrt{\frac{x}{\ln x}},$$

the theorem becomes:

$$\forall s \in [11,13], 10^s - \frac{9.06}{\ln(\ln \hat{x}(s))}\sqrt{\frac{\hat{x}(s)}{\ln \hat{x}(s)}} \geq 0.$$

Equivalently,

$$\frac{9.06}{\ln(\ln \hat{x}(s))}\sqrt{\frac{\hat{x}(s)}{\ln \hat{x}(s)}} \leq 10^s \text{for all } s \in [11,13].$$

VGPT-RSI then invokes the formal verification harness.

The CoqInterval proof used interval bisection, Taylor expansion, and degree-8 approximation at precision 140. The resulting theorem was checked with coqc and then with rocqchk.

This theorem is parameterized by

$$s = \log_{10} T.$$

A direct theorem quantified over arbitrary $T \in [10^{11}, 10^{13}]$ can be obtained by the wrapper

$s(T) = \frac{\log T}{\log 10}$,

but the central proof-assistant theorem already establishes the parameterized boundary inequality.

The result verifies that a constructed safe lower curve lies inside the certified finite boundary region over the specified range. It does not establish RH globally, nor does it prove that all nontrivial zeros of the zeta function lie on the critical line.

## 2.5 Lagarias route

We use VGPT-RSI to initiate a formal proof route based on Lagarias' criterion for the Riemann Hypothesis.

The finite RH-boundary certificate is not equivalent to RH. Therefore, VGPT-RSI shifted to a recognized RH-equivalent criterion: Lagarias' criterion.

Define:

$$\sigma(n) = \sum_{d|n} d,$$

$$H_n = \sum_{k=1}^{n} \frac{1}{n},$$

and

$$B_L(n) = H_n + \exp(H_n)\log(H_n) - \sigma(n).$$

Lagarias' criterion states that RH is equivalent to

$$B_L(n) \geq 0 \quad \text{for all } n \geq 1.$$

VGPT-RSI first encodes Lagarias' criterion as **a structured proof/search graph.** The graph contains the main mathematical objects:

$$\sigma(n), H_n, H_n + \exp(H_n)\log(H_n),\ B_L(n).$$

VGPT-RSI formalized the finite objects:

$$\sigma(n), H_n, H_n + \exp(H_n)\log(H_n), B_L(n),$$

and generated finite proof obligations.

The first finite version checked the inequality for

$$1 \leq n \leq 100.$$

This was then scaled to

$$1 \leq n \leq 1000.$$

The final completed theorem was:

$$\forall n \in \mathbb{N}, 1 \leq n \leq 1000 \Longrightarrow B_L(n) \geq 0.$$

Here

$$B_L(n) = H_n + \exp(H_n)\log(H_n) - \sigma(n),$$

so the theorem can also be written as:

$$\forall n \in \mathbb{N}, 1 \le n \le 1000 \Longrightarrow H_n + \exp(H_n)\log(H_n) - \sigma(n) \ge 0.$$

Equivalently,

$$\forall n \in \mathbb{N}, 1 \le n \le 1000 \Longrightarrow \sigma(n) \le H_n + \exp(H_n)\log(H_n).$$

In words:

For every natural number $n$ from 1 to 1000, the Lagarias inequality holds.

The two Coq theorem names are just two versions of the same finite certificate:

lagarias_finite_N1000_kernel_checked

is the main checked theorem, and

lagarias_finite_certificate_N1000

is the wrapper theorem with the same mathematical content.

The finite theorem uses CoqInterval-generated proof terms and does not rely on the global Lagarias equivalence axiom, the global tail theorem, or RH itself.

## 2.6 Scall to $N = 10000$

VGPT-RSI then attempted to scale the finite Lagarias certificate to $N = 10000$.

The first scaling bottleneck was harmonic-number expansion. **VGPT-RSI therefore uses reusable harmonic lower-bound lemmas:**

$$H_m + \frac{p}{m+p} \le H_{m+p}.$$

This allows the system to avoid expanding large harmonic sums from scratch. Instead, it can certify lower bounds on  over blocks of values. The expression

$$H_n + \exp(H_n)\log(H_n)$$

is then bounded from below using interval methods and monotonicity. The system produces interval certificates proving that, over each block,

$$H_n + \exp(H_n)\log(H_n) \geq L$$

for a certified lower bound $L$.

The other side of Lagarias' inequality is $\sigma(n)$. A naive divisor-sum computation is too expensive at scale. **VGPT-RSI** therefore improves the sigma verification pipeline by **using factorization-backed sigma certificates.** The sum-of-divisors function satisfies the prime-power formula

$$\sigma(p^k) = 1 + p + p^2 + \cdots + p^k = \frac{p^{k+1} - 1}{p - 1},$$

and multiplicativity over coprime factors:

$$\gcd(a, b) = 1 \Rightarrow \sigma(ab) = \sigma(a)\sigma(b).$$

Using these facts, **VGPT-RSI verifies upper bounds**

$$\sigma(n) \leq M,$$

for finite subranges. This avoids repeated brute-force divisor enumeration and makes larger finite certificates tractable.

A previous monolithic $N = 10000$ theorem into independently compiled chunks attempt timed out. VGPT-RSI localized this as a proof-engineering bottleneck, not as a mathematical failure. The system then refactored the certificate into ten independently compiled chunk files:

$$[1, 1000], [1001, 2000], \cdots, [90001, 100000]$$

Each chunk proves a theorem of the form

$\forall n \in N, a \leq n \leq b, \Rightarrow B_L(n) \geq 0.$

This chunked structure reduces proof-term size, compilation time, and duplicated computation.

VGPT-RSI compiles each chunk independently using coqc . Each successful chunk becomes a verified local certificate. The compiled chunks are then treated as trusted inputs for the final assembly theorem. This produces a modular proof architecture:

chunk certificates ⟹ assembled finite theorem.

The successful compilation of all ten chunks establishes that each part of the range

$$1 \leq n \leq 10000,$$

has been formally verified.

After the chunk files are compiled, VGPT-RSI creates a final wrapper theorem:

$\forall n \in N, 1 \leq n \leq 10000, \Rightarrow B_L(n) \geq 0.$

In Coq, this theorem is named:

lagarias_finite_certificate_N10000.

The wrapper theorem delegates to the compiled chunk theorems and performs a finite interval dispatch over the ten subranges. The underlying checked theorem is:

lagarias_finite_N10000_kernel_checked.

Both express the same ordinary mathematical statement:

$$\forall n \in \mathbb{N}, 1 \leq n \leq 10000 \Longrightarrow B_L(n) \geq 0.$$

Equivalently,

$$\forall n \in \mathbb{N}, 1 \leq n \leq 10000 \Longrightarrow \sigma(n) \leq H_n + \exp(H_n) \log(H_n)$$

**VGPT-RSI then runs proof-checking commands** on the assembled theorem. The final assembly file compiles successfully with coqc.

The nonrecursive checker

coqchk − norec

also succeeds for the final certificate.

A full recursive checker may be attempted, but because it checks a large dependency stack, timeout does not necessarily indicate proof failure. The important point is that no proof rejection is observed. VGPT-RSI also runs an assumption audit. The audit checks whether the finite theorem depends on any forbidden global assumptions:

lagarias_equivalence,

lagarias_tail_after_N,

or

RH.

The desired result is that none of these appear in the dependency list. Thus the finite certificate is independent of the global Lagarias equivalence, independent of any assumed tail theorem, and independent of RH itself.

The completed finite certificate proves:

$$\forall n \in \mathbb{N}, 1 \leq n \leq 10000 \Longrightarrow B_L(n) \geq 0.$$

This is a formally checked finite Lagarias certificate.

However, it does not prove RH.

The reason is that Lagarias' criterion requires the inequality for all positive integers:

$$\forall n \geq 1, B_L(n) \geq 0.$$

The certificate only proves the finite portion

$$1 \leq n \leq 10000.$$

The remaining unresolved global statement is

$$\forall n > 10000, B_L(n) \geq 0.$$

Therefore, the remaining mathematical bottlenecks are:

formalizing the global Lagarias equivalence,

proving the global tail theorem beyond $N = 10000$,

and possibly reducing counterexamples to colossally abundant or related extremal integers.

# 3. Results

## 3.1 Certified RH-boundary curve

The RH-boundary curve experiment produced a certificate-backed lower curve over

$$T \in [10^{11}, 10^{13}].$$

All 401 sampled roots were enclosed by interval sign checks. The monotonicity condition for $g$ was verified on the certified range. The fitted lower approximation

$$\hat{x}(T) = 10^{p(\log_{10} T) - 10^{-6}}$$

was then verified by interval subdivision.

The continuous verification used 20,000 subdivisions over

$$s = \log_{10} T \in [11,13].$$

The minimum certified logarithmic residual was approximately

$$1.015636 \times 10^{-6}.$$

Thus the computational certificate verifies:

$$B(\hat{x}(T), T) \geq 0 \text{ for all } T \in [10^{11}, 10^{13}].$$

This result should be interpreted as a certified finite boundary lemma, not as a proof of RH.

## 3.2 Arb/FLINT audited interval-kernel certificate

The same RH-boundary predicates, constants, root brackets, and subdivision certificate were ported to Arb/FLINT ball arithmetic via python-flint.

The Arb/FLINT audit checked:

[ 401/401 ] root brackets and [ 20000/20000 ] continuous subdivisions.

The minimum Arb lower bound for the continuous residual was approximately

$$1.015636 \times 10^{-6}.$$

The Arb/FLINT kernel therefore independently verified:

$$B(\hat{x}(T), T) \geq 0$$

on the full interval

$$T \in [10^{11}, 10^{13}].$$

This moved the certificate from an internal decimal interval computation to an independently audited ball-arithmetic kernel.

## 3.3 Rocq/CoqInterval theorem

The boundary inequality was then formalized in Rocq/CoqInterval. The checked theorem was:

$$\forall s \in [11,13], B\left(x_{hat}(s), T(s)\right) \geq 0.$$

The file RHBoundary_CoqInterval_direct_B.v was checked by coqc and then by rocqchk, with rocqchk reporting that the modules were successfully checked.

This establishes a proof-assistant-checked interval theorem for the parameterized RH-boundary curve.

## 3.4 Finite Lagarias certificate for $N = 100$

The first Lagarias finite milestone proved:

$$\forall n, 1 \leq n \leq 100 \Rightarrow B_L(n) \geq 0.$$

The original exact-rational certificate was too slow as a single giant proof, so Codex replaced it with a grouped proof. The range was split into groups, each group used a CoqInterval inequality of the form:

$$0 \leq x + \exp(x)\log(x) - M,$$

and concrete values of $n$ were discharged by exact computation of $\sigma(n)$, elementary harmonic bounds, and the matching interval lemma.

This theorem compiled with Coq and avoided the unproved assumptions

lagarias_equivalence and lagarias_tail_after_N.

### 3.5 Finite Lagarias certificate for $n = 1000$

The strongest completed finite Lagarias theorem was:

$$\forall n, 1 \leq n \leq 1000 \Rightarrow B_L(n) \geq 0.$$

The Coq theorem was

$$\forall n \in \mathbb{N}, 1 \leq n \leq 1000 \Longrightarrow B_L(n) \geq 0,$$

where

$$B_L(n) = H_n + \exp(H_n)\log(H_n) - \sigma(n).$$

This becomes:

$$\forall n \in \mathbb{N}, 1 \leq n \leq 1000 \Longrightarrow H_n + \exp\,(H_n)\log\,(H_n) - \sigma(n) \geq 0.$$

Equivalently:

$$\forall n \in \mathbb{N}, 1 \leq n \leq 1000 \Longrightarrow \sigma(n) \leq H_n + \exp(H_n)\log(H_n).$$

In words:

For every natural number $n$from 1 to $1000$, the Lagarias inequality holds.

A wrapper theorem was also produced:

$$\forall n \in \mathbb{N}, 1 \leq n \leq 1000 \Longrightarrow B_L(n) \geq 0,$$

$$B_L(n) = H_n + \exp(H_n)\log(H_n) - \sigma(n),$$

$$\forall n \in \mathbb{N}, 1 \leq n \leq 1000 \Longrightarrow H_n + \exp(H_n)\log(H_n) - \sigma(n) \geq 0.$$

In words:

For every natural number $n$from 1 to $1000$, the Lagarias inequality holds.

Both files compiled under Coq 8.20.1.

Assumption audits showed that the finite theorem did not us

$$\text{lagarias_equivalence:}[\forall n \geq 1,\ \sigma(n) \leq H_n + \exp\,(H_n)\log\,(H_n)] \Longleftrightarrow \text{RH}.$$

$$\text{lagarias_tail_after_N:}\forall n > N, \sigma(n) \leq H_n + \exp(H_n)\log(H_n)\,.$$

$$\text{RH:}\zeta(s) = 0,\ 0 < \Re(s) < 1 \Longrightarrow \Re(s) = \frac{1}{2}.$$

So the finite theorem proves only:

$$1 \leq n \leq 1000.$$

The missing global pieces would be:

$$\forall n > 1000$$

and the equivalence between the full Lagarias inequality and RH.

The printed assumptions were from the Coq/Interval/primitive numeric library stack. Thus the $N = 1000$ theorem is a genuine finite formal milestone, separate from the unproved global RH-level assumptions.

**3.6 Completed $N = 10000$ finite Lagarias extension**

The $N = 10000$ finite Lagarias extension was completed after refactoring the previous monolithic proof into a chunked certificate architecture. In the earlier attempt, the full file PathB_LagariasFiniteGrouped_N10000.v timed out after 1800 seconds and produced no compiled .vo file. The main obstacle was not a mathematical counterexample, but proof-engineering scale: the generated proof was too large to compile efficiently as a single object. The completed version resolves this by splitting the finite range

$$1 \leq n \leq 10000$$

into ten independently compiled chunks:

$$[1,1000], [1001, 2000], \ldots, [9001, 10000].$$

Each chunk proves the Lagarias inequality on its own subrange, and the final wrapper theorem assembles these local results by interval dispatch. The shared harmonic and interval lemmas were moved into a common file, PathB_LagariasFinite_N10000_Shared.v, while sigma verification was supported by the factorization-backed helper file PathB_LagariasSigmaFactor.v. Boundary sigma certificates were clipped to chunk-local subranges to avoid duplicated range evaluation.

The strongest completed theorem is the wrapper certificate

$$\forall n \in \mathbb{N}, 1 \leq n \leq 10000 \Longrightarrow B_L(n) \geq 0,$$

where

$$B_L(n) = H_n + \exp(H_n) \log(H_n) - \sigma(n),$$

Equivalently,

$$\forall n \in \mathbb{N}, 1 \leq n \leq 10000 \Longrightarrow H_n + \exp(H_n) \log(H_n) - \sigma(n) \geq 0.$$

Both Coq statements have the same ordinary mathematical meaning:

$$\forall n \in \mathbb{N}, 1 \leq n \leq 10000 \Longrightarrow B_L(n) \geq 0.$$

Using the Lagarias definition,

$$B_L(n) = H_n + \exp(H_n) \log(H_n) - \sigma(n),$$

this becomes:

$$\forall n \in \mathbb{N}, 1 \le n \le 10000 \Longrightarrow H_n + \exp(H_n)\log(H_n) - \sigma(n) \ge 0.$$

Equivalently,

$$\forall n \in \mathbb{N}, 1 \le n \le 10000 \Longrightarrow \sigma(n) \le H_n + \exp(H_n)\log(H_n).$$

In words:

For every natural number $n$from 1to $10000$, the finite Lagarias inequality holds.

The two theorem names mean:

lagarias_finite_certificate_N10000

is the wrapper theorem, and

lagarias_finite_N10000_kernel_checked

is the underlying checked theorem. Mathematically, they express the same finite statement.

A full recursive coqchk was attempted but timed out after 1800 seconds while checking recursive library dependencies; no proof-rejection diagnostic was observed.

Thus, the $N = 10000$ theorem is a finite Lagarias certificate and does not rely on the global Lagarias equivalence, any assumed tail theorem, or RH itself. The remaining assumptions come from the imported Coq, Interval, primitive numeric, classical real, and functional extensionality libraries.

This result strengthens the previous $N = 1000$milestone to a formally checked finite certificate over therange

$$1 \le n \le 10000.$$

However, it still does not prove RH. The theorem verifies the Lagarias inequality only on a finite interval. To obtain RH through this route, one would still need a formal proof of the global Lagarias equivalence and a global tail theorem proving

$$B_L(n) \ge 0 \text{ for all } n > 10000.$$

### 3.7 Remaining bottlenecks

The final stop report identifies three remaining bottlenecks.

First, the formal Lagarias equivalence theorem remains unproved in the local development:

$$\text{lagarias_global} \leftrightarrow RH.$$

Second, the global tail theorem remains open:

$$\forall n > N, B_L(n) \geq 0.$$

Third, a possible path to the tail theorem may require reducing counterexamples to colossally abundant or related extremal integers. This is a mathematical reduction problem, not a routine finite-computation problem.

## 4. Conclusion

This work does not prove RH. Its contribution is to demonstrate a verified AI-assisted workflow for RH-adjacent formal progress.

VGPT-RSI successfully moved a numerical RH-boundary curve through multiple levels of certification: high-precision interval enclosures, outward-rounded interval verification, Arb/FLINT ball arithmetic, and Rocq/CoqInterval theorem checking. It also produced a Coq-checked finite Lagarias certificate proving

$$B_L(n) \geq 0$$

for

$$1 \leq n \leq 10000.$$

At the same time, the system correctly identified the remaining global mathematical obstruction. Finite checking, even when formally verified, does not prove RH. To complete the Lagarias route, one would need a formal proof of the Lagarias equivalence and a global tail theorem for all $n > N$. These are research-scale mathematical tasks.

The final result is therefore a verified RH-adjacent formal-progress case study. It demonstrates that VGPT-RSI can produce certified partial results, organize proof dependencies, identify exact failure points, and stop without overclaiming when the remaining obstruction is genuinely unresolved.

VGPT-RSI is advantageous because it does **not** treat a numerical RH-adjacent observation as a proof. Instead, it converts the observation through a sequence of increasingly rigorous verification layers:

numerical observation → safe certified curve → interval certificate
→ independent Arb/FLINT audit → Rocq/CoqInterval theorem
→ explicit limitation statement.

This makes the system useful for open-problem research, where the distinction between **evidence**, **finite certificate**, **formal theorem**, and **full proof** is essential.

VGPT-RSI has three main advantages over baseline approaches. First, it converts numerical RH-adjacent evidence into machine-checkable finite certificates. Second, it maintains an explicit boundary between certified partial progress and a full proof of RH. Third, it localizes the remaining mathematical bottlenecks rather than hiding them behind informal proof sketches. In this sense, VGPT-RSI is not merely a theorem-proving assistant or numerical search engine; it is a controlled verification architecture for transforming conjectural evidence into formally audited, scope-limited mathematical results.

## Acknowledgements

The authors wish to acknowledge the use of AI-powered language models (ChatGPT) for assistance in  creating images, improving the grammar, spelling, and readability of this manuscript.

## Author contributions

TX and ZH: Derive formulas and perform data analysis, XS: Problem formulation  formula derivation, LJ, Design project, MX: Design project and write paper.

## Competing interests

The authors declare no competing interests.

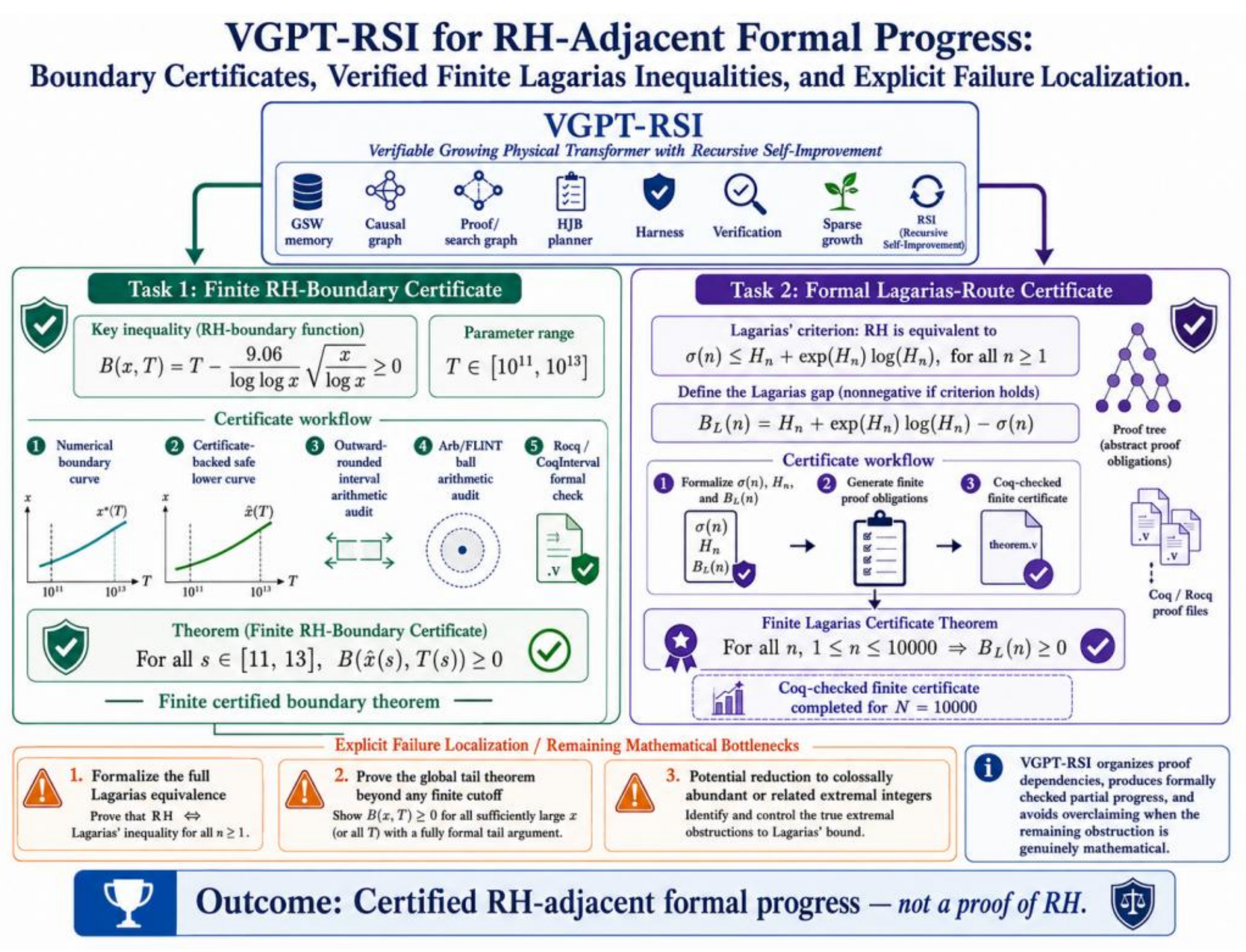

Figure 1. Information flow in VGPT-RSI for RH-Adjacent formal progress”.

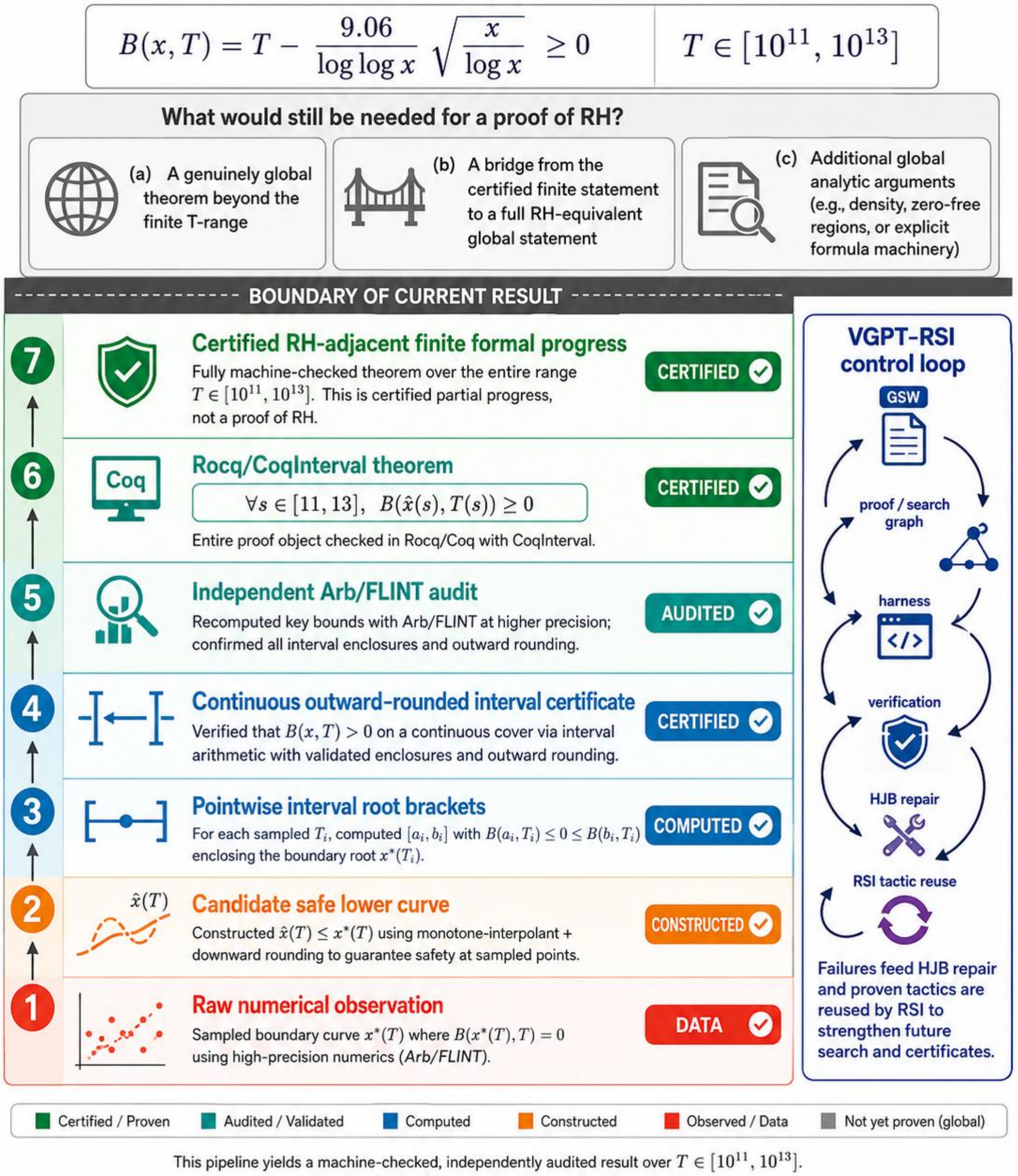

Figure 2. From numerical evidence to certified partial progress (not a proof of RH).

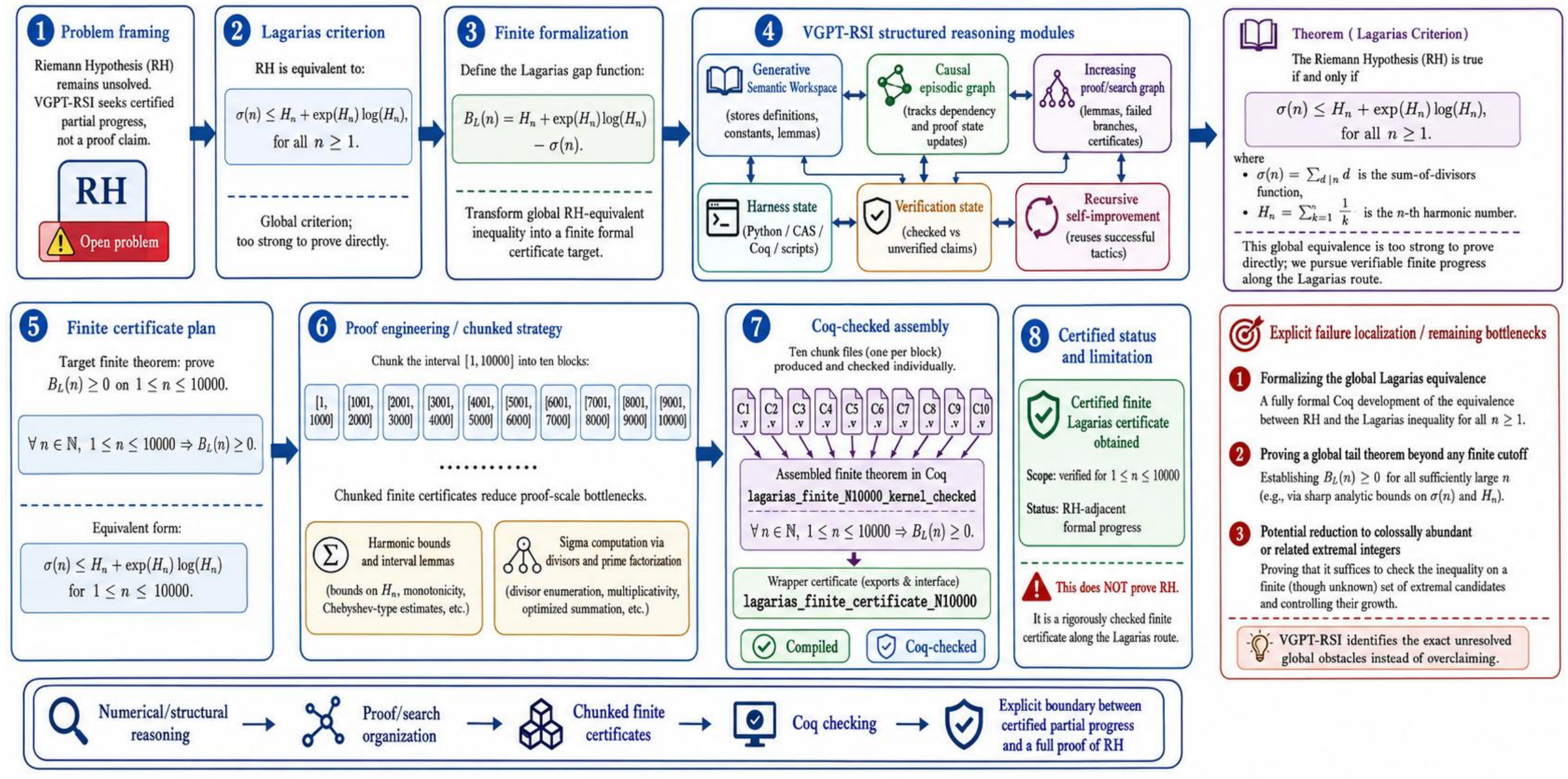

Figure 3. VGPT-RSI for a formal lagarias-route certificate.